# The Application of a Dendritic Cell Algorithm to a Robotic Classifier

Robert Oates Julie
Greensmith Uwe
Aickelin Jonathan
Garibaldi Graham
Kendall

The University of Nottingham
{ rxo, jqg, uxa, jmg, gxk } @cs.nott.ac.uk,
http://www.asap.cs.nott.ac.uk

**Abstract.** The dendritic cell algorithm is an immune-inspired technique for processing time-dependant data. Here we propose it as a possible solution for a robotic classification problem. The dendritic cell algorithm is implemented on a real robot and an investigation is performed into the effects of varying the migration threshold median for the cell population. The algorithm performs well on a classification task with very little tuning. Ways of extending the implementation to allow it to be used as a classifier within the field of robotic security are suggested.

## 1 Introduction

Technologies and protocols designed to enforce security are now pervasive in society. Most houses now have burglar alarms, CCTV is common-place in towns and cities and the private security industry is estimated to provide products and services up to the value of £4 billion in the UK alone [1]. It is possible to group most existing solutions as either 'manned guarding' or static-sensor networks.

Manned guarding (bouncers, private security guards etc.) is a popular technique for providing additional security to buildings containing expensive or sensitive items. Human security systems are difficult to pre-empt and can adapt to new circumstances. However, human performance varies greatly and is heavily reliant on rest periods. People are also susceptible to prejudices and preferences depending on gender, race and age. Guarding is a potentially hazardous occupation as it places an individual between a criminal and their goal.

Static sensor networks (CCTV, standard burglar alarms etc.), can be stored and replayed as and when required. They do not require rest and react predictably to all situations. If damaged, static sensors are easy to replace and in systems with centralised data storage, evidence is not compromised. However, criminals can plan around static sensors; as they can be obscured and cannot negotiate obstacles. Static sensor networks cannot effectively use short-range sensor-types, unless deployed in bottle-necks, such as entry and exit points. The limiting factor for many static sensor networks is the volume of information generated. Very few sensors can be monitored by an individual effectively. Tickner et al. estimated that the number of feeds that a single operator can effectively monitor is approximately 16, with the detection rate falling from 83% for a four camera system, to 64% for a 16 camera system, [2].

Robotic systems have many properties to make them a useful tool for security applications. They have the advantages of static sensor networks and are capable of moving around obstructions to gain a better line of sight. Short range sensors are more effective when mounted on a robot, as the sensor can be taken to the target. Whilst it is possible for an automated sentry to become predictable, intelligent routing algorithms could make evasion challenging. The key disadvantage of a robotic system is the volume of data. The camera mounted to the front of a robot is likely to be even more difficult to monitor than a static camera, as both the background, and items of interest will be moving on the screen. This disadvantage could potentially be overcome if the robots could autonomously recognise events of interest and report them to the operator.

Artificial immune systems (AIS) have had numerous successes in the field of anomaly detection. A newly developed AIS algorithm, the dendritic cell algorithm (DCA), is a promising technique for the processing of time-dependant data [3]. The DCA is based on recent developments in immunology regarding the role of dendritic cells (DCs), as a major control component within the immune system. The DCA is based on an abstraction of DC behaviour and performs fusion of data from disparate sources. Successful applications of the DCA have focussed on solving intrusion detection problems in computer security, a field which shares properties with problems in both robotics and physical security.

The potential benefits of applying the DCA to a robotic security solution are numerous. It is hoped that the DCA will have a resilience to the noise associated with real-world signals due to its ability to fuse information from disparate sources via a population of artificial cells. The aim of this investigation is to explore the applicability of the DCA to a robotic system. In section 2 we present work relevant to the areas of security robotics and the DCA. In section 3 the implementation of a general robotic DCA is discussed. Section 4 outlines an investigation into the effects on the performance of altering the dendritic cell migration threshold when applied to a trivial robotic classification problem. The results of this investigation are presented, analysed and discussed in sections 4.1 to 4.6. In section 5 conclusions are drawn about the applicability of the algorithm to robotic security and possible extensions of this work are outlined.

## 2 Related Work

### 2.1 Robotic Systems

Developing a robotic system is a demanding task as robust, real-time control is difficult to achieve. Brooks' "subsumption architecture" (first proposed in [4]), has been shown to be an effective way of designing robotic control systems [5]. Such architectures rely on the development of a family of simplistic

"behavioural modules" that interact to produce more complex behaviour. For example, the complex behaviour of wandering through a dynamic environment, without hitting obstacles can be achieved through the interaction of two simplistic behavioural modules. Figure 1 illustrates a simple subsumption architecture. The lower priority behaviour simply moves the robot forwards at a constant velocity. In the event of a higher priority behaviour detecting an obstacle, it can subsume the output of the low priority behaviour and steer the robot away or, in emergencies, stop. The interaction between the two modules ensures that the robot is always moving when possible, without hitting obstacles.

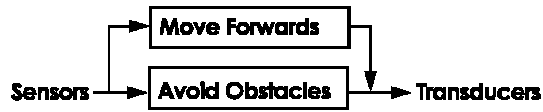

Fig. 1. A simple subsumption architecture for implementing a wandering behaviour

### 2.2 Autonomous Security Systems

Whilst the robotic security problem is yet to be rigorously formalised, an architecture using robots as autonomous scouts which report 'interesting' events to a human operator has precedent [6][7][8]. Using this approach, the robotic security problem can be viewed as two, well-researched problems: path planning and classification. Massios et al. [9] define patrol route planning as an optimisation problem, minimising the probability of missing a "relevant event".

The classification problem is the discrimination of important events from normal events. The "mobile detection assessment and response system" [8] is an American military project aimed at producing a collection of robots for interior and exterior security. These systems use basic motion-detection algorithms on data from an on-board camera [7]. The movement detection algorithm is simplified by keeping the robot base stationary during classification. When movement is detected other sensors are employed in conjunction with the camera to assess if the observed object is human or not. In [6], a more intelligent classification technique is proposed using colour analysis and clustering to compare a room's current state with its previously observed state. This algorithm has applications for identifying erroneous objects, e.g. unattended luggage, and recognising the theft of objects that were present in the test image.

### 2.3 The Dendritic Cell Algorithm

The DCA was conceptualised and developed by Greensmith et al. [10]. The algorithm is based on the behaviour of DCs, which are the antigen presenting cells of the immune system. DCs are natural anomaly detectors and data fusion agents, responsible for controlling and directing appropriate immune responses. The fusion of 'signals' across a population of DCs and the asynchronous correlation of signals with 'antigen' provides the basis of the DCA's classification. DCs exist in one of three states, immature, semi-mature and mature. Immature DCs perform signal fusion and process antigen. Semi-mature and mature DCs present antigen with a context value derived from the fused signals. Antigen presented by semi-mature DCs are 'normal' and the antigen presented by mature DCs are 'anomalous'. The biological theory is beyond the scope of this paper, but the interested reader can refer to [10] and [11] for the relevant immunological details. A formal description of the DCA is provided in [3].

The DCA is a population-based algorithm, with each agent in the system represented as a cell. Each cell can collect data items to classify forming antigen for use within the DCA. The DCA used in this paper relies on a '3-signal' model where three categories of input signal are used to produce three output signals. Signals and antigen are read into a signal matrix and antigen vectors. Antigen is sampled by DCs and removed from the tissue antigen vector and transferred to the DC's own antigen storage facility. Once antigen is sampled, the DC copies the values of the tissue signal matrix to its own signal matrix. These values are processed by the DC during the update to form cumulative output signal values. Equation 1 is the function used to process the signals, where o are output signals, S are input signals, i is the number of output signals, j is the number of input signals and $W_{ij}$ is the weight used for $o_i$ and $S_j$.

$$O_i = \sum_{j=1}^{3} W_{ij} S_j \quad \forall i \tag{1}$$

Input and output signals are termed after their biological counterparts:

PAMPs ($S_1$): A signature of abnormal behaviour e.g. number of errors per second. This signal is proportional to confidence of abnormality.

Danger Signal ($S_2$): A measure of an attribute which increases in value to indicate an abnormality e.g. an increase in the rate of a monitored attribute. Low values of this signal may not be anomalous, giving a high value a moderate confidence of indicating abnormality.

Safe Signal ($S_3$): A measure which increases value in conjunction with observed normal behaviour e.g. a high value of $S_3$ is generated if the standard deviation of a monitored attribute is low. This is a confident indicator of normal, predictable or steady-state system behaviour. This signal is used to counteract the effects of PAMPs and danger signals and is assigned a negative weight in the weighted sum.

CSM ($o_1$): The costimulatory signal which is increased as a result of high values of all input signals. This value is used to limit the duration spent by DCs in the data sampling stage.

IL-10 ($o_2$): This value is increased upon the receipt of the safe signal alone.
IL-12 ($o_3$): This value is increased upon the receipt of PAMP and danger signals, and is decreased by the safe signal.

The processing of signals and antigen is distributed across the DC population to correlate disparate data sources to perform the classification of the algorithm. The DCA does not perform antigen pattern matching, unlike other AIS algorithms which perform antigen classification through analysis of the the structure of an antigen. Instead, the signals received by a DC during its antigen collection phase are used to derive an antigen context which is used to perform the basis of classification. This algorithm can be applied to problems where multiple antigens of identical structure i.e. antigens of the same type, are to be classified, such as the classification of anomalous processes [12].

Each DC is randomly assigned a migration threshold value which is compared against the cumulative $o1$ value. The details of the migration threshold value generation for this experiment can be found in section 4.4. If the value of $o1$ exceeds the migration threshold, the DC is removed from data sampling and enters the maturation stage. At this point the values for output signals $o2$ and $o3$ are assessed. If $o2 > o3$, the DC is termed 'semi-mature'. Antigen 'presented' by a semi-mature cell is assigned a context value of 0. Conversely, if $o2 < o3$ the cell is termed 'mature' and antigen presented by this cell is assigned a context value of 1. Once the DC has presented its antigen-plus-context values, it is reset and returned to the DC population. Data sampling by DCs continues for the duration of the experiment, or until a specified stopping condition is met.

After a specified number of antigen are presented by the DCs, analysis is performed. As mentioned, antigen do not have unique values representing their structure, with antigen of identical values termed as a 'type'. The MCAV coefficient (mature context antigen value) is calculated as the fraction of antigen presented in the mature context, per type of antigen. MCAVs close to 1 indicate that a type of antigen is potentially anomalous. A threshold is applied to the MCAV values to discriminate between anomalous and normal types of antigen.

Thus far, the majority of problems presented to the DCA are related to computer security, specifically the detection of port scans [3] it is also applied to a static machine learning dataset [10] and the detection of intrusions in sensor networks[13]. Work performed by Greensmith et al. [12] has indicated that the DCA performs well for time-dependent real-valued data, such as that seen in robotics applications. Following the interesting ideas proposed in [14], it is fathomable that the DCA, also based on innate immunity, could be incorporated into the field of mobile robotics. As the DCA has a history of good performance for illegal scan detection in computer security, it may be a useful algorithm for the purpose of physical robotic security applications.

## 3 The Robotic Dendritic Cell Algorithm

The DCA is applied to a general robotics problem to support the suitability of the algorithm for mobile robotic security. The platform used for this investigation is a

Pioneer 3DX. This robotic system has a broad variety of sensors, including a laser range finder (LRF), an array of sonar sensors and a pan-tilt-zoom camera. On-board processing is performed using an 850MHz Pentium III processor running Debian Linux, (kernel version 2.6.10). The manufacturer's "Aria" library is used to control the device. The Aria control system is an object-orientated (C++) library which is structured to support the implementation of subsumption control architectures. All compilation was carried out using g++ version 4.0.2.

The robotic DCA is implemented as a stand-alone behavioural module for compatibility with a subsumption architecture. Figure 2 illustrates the architecture which implements a simultaneous wandering and DCA classifying behaviour. This extension of the Aria library's 'wander' architecture has an additional module for image processing and an additional module for executing the DCA. By making these additions part of the subsumption architecture, the fundamental behaviour of moving around safely within the environment can be prioritised above all other actions. In addition to the wandering and classifying behaviour, there is also a tele-operation, (remotely controlling the robot from a networked machine) and classifying behaviour. The DCA module outputs MCAV coefficients (as described in section 2.3), approximately once per second.

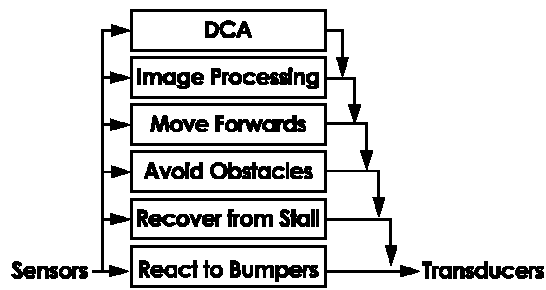

Fig. 2. The subsumption architecture used to implement the robotic DCA

The DCA used on the robot is a streamlined version of the algorithm which does not require any additional software libraries, unlike the implementation used in [12]. Verification of the streamlined implementation's functionality against that of the original DCA has been achieved. This was performed by attaching 'virtual' signals and antigen to the inputs of the module and processing the data used in [12]. The signal weightings specified for the anomaly detection algorithm in [12] were used for all experiments.

## 4 Experimental Validation

It is thought that the DCA is capable of processing real-time sensor data. It is further hypothesised that the migration threshold will have a noticeable effect on the false positive rate for this classification task. The following experiment is designed to test these ideas.

### 4.1 Experimentation

This experiment uses the "simultaneous wander and classify" behaviour discussed in section 3. The DCA classifies its current location as either 'anomalous' or 'normal' from the application-specific input signals. For this simple test, pink coloured objects with a height less than 330mm are considered anomalous whilst other obstacles are considered to be normal. The colour pink is used as it is easily distinguished from other objects within the robot's environment. A height of 330mm is used as objects below that height are unobservable by the LRF's planar field of view (FOV), but can still be detected by the sonar sensors' conic FOV. This means objects classified as 'pink', detectable by the sonar but not detectable by the laser, are classified as anomalous.

The starting conditions for the experiments are illustrated by figure 3. Obstacle A is a pink cylinder, with a height less than 330mm which is an example of an anomalous object. Obstacle B is a pink cylinder, with a height greater than 330mm which is an example of a normal object. It is expected that the DCA will not react to the taller cylinder as the algorithm will prevent full maturation of the cells. By maintaining the starting position of the robot and the positions of the obstacles, it is possible to calculate the ideal classification for all points within the enclosure. The error between the theoretical response and the algorithm output can then be used as a metric to assess the performance of the algorithm.

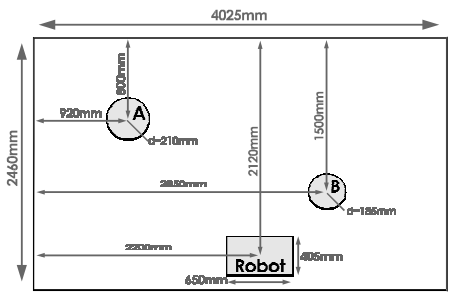

Fig. 3. The starting conditions for each experiment. Cylinder A is the 'dangerous' obstacle, cylinder B is the 'safe' obstacle.

### 4.2 Signal Sources

As described in section 2.3, three signals are used as inputs to the DCA inclusive of a safe signal, a danger signal and a PAMP signal. The former acts to suppress the full maturation of the dendritic cells, whilst the other two stimulate the maturation. All signals contribute to the migration of the cells.

The PAMP is sourced from the image processing module. The input from the camera is transferred into the HSV, (Hue, Saturation, Value) colour space and the histogram back-projection algorithm is applied to the data [15]. The back-projection algorithm uses a single training image to identify the colour properties of an object of interest. All pixel groups within the image that share the same statistical properties are identified and contours are drawn around those clusters. The final output from the image processing library is the area of the largest region which matches the properties of the test image. Intel's "OpenCV" library was used to perform all image processing. The output from the image processing module is scaled down before being used as the PAMP signal. The scaling factor used was calculated from test data generated by a seven minute random walk around the pen.

The LRF is used as the source for the safe signal so objects taller than 330mm will produce high values of the inhibitory signal. The FOV of the LRF extends from -90° to +90° (where 0° is directly in front of the robot). A 44° FOV is used, ranging from -22° to +22°. A narrow FOV reduces the risk of erroneous classification from walls. The distance to the closest object within the safe FOV is returned to the signal processor. The signal processor calculates the magnitude of the safe signal. This is performed using a look up table which relates distance to signal strength. For values that lie between those specified in the look up table, linear interpolation is used to calculate the signal strength. The values used are given in Table 1.

Table 1. Object Distance and Signal Strength for Ranged Sensors

| Distance (mm) | Safe Signal Strength |
|---|---|
| 0 | 100 |
| 300 | 90 |
| 600 | 50 |
| 900 | 20 |
| 1200 | 0 |

The danger signal is sourced from the sonar array which has a 360° FOV. The danger signal FOV coincides with the safe signal FOV. The same look up table (see Table 1) used to normalise the laser output for the safe signal is used to normalise the sonar output for the danger signal.

### 4.3 Antigen Source

In a practical robotic security solution, the antigen could be a vector based on the estimated location of the anomalous situation. Object-based approaches for antigen generation within a robot system have been put forward by by Krautmacher et al. in [16]. For this simple implementation antigen is an integer number which uniquely identifies a segment of the test pen. This encapsulates a small range of positions and orientations of the robot. The actual position and orientation of the robot is estimated using a 'dead reckoning' algorithm. Dead reckoning estimates the position and orientation of the robot from encoders mounted on the wheels, the fixed starting position of the robot and the diameter of the tyres. The antigen generated enumerates a 300mm grid square within the pen and a 30° segment within that square. Generating antigen based on the current location of the robot is more practical than object-based antigen, which requires a deeper knowledge of the environment to compute.

As the antigen is generated based on a specific robot location, it is possible that an ineffective amount of antigen will be generated. One solution for this is to add multiple copies of each antigen to the DCA environment as suggested in [17]. A novel extension to the DCA for this application is an antigen multiplication function. This function adds varying amounts of each antigen depending on the speed of the robot. Areas passed through slowly are made to contribute more antigen than areas passed through quickly. This is done because areas passed through quickly contribute less signal to the DCA environment, as less time is physically spent within that area. The weighting function is given in equation 2.

$$W(v, \dot{\theta}) = 75\left(1 - \left|\frac{v}{v_{max}}\right|\right) + 1 + 25\left(1 - \left|\frac{\dot{\theta}}{\dot{\theta}_{max}}\right|\right) + 1 \qquad (2)$$

In equation 2 v is the velocity of the robot, $\dot{\theta}$ is the rotational velocity of the robot and W is the amount of antigen added to the environment.

The smallest amount of antigen that can be added is 2, when the robot is at maximum velocity and maximum rotational velocity. The maximum amount of antigen that can be added is 102, when the robot is totally stationary.

A simple program written in Java calculates the theoretical MCAVs for every antigen, from the properties of the test pen. This is an unrealistic mathematical model of the experiment but provides a way of analysing the true and false positive rates for each run of the experiment.

4.4 Experiment Parameters

Each run of the experiment allowed the robot to wander around the test pen for ten minutes. The classification experiment was repeated three times for each value of migration threshold median. The experiments used the migration medians 15, 30, 60, 120 and 240. The range of allowed values is ±50% of the migration median in each case. Each DC was assigned a random migration threshold within the specified range, using an equi-probable distribution.

A naming convention is used referring to the first experiment as M15, the second as M30 etc. A threshold of 0.6 is applied to the MCAV values from the algorithm. Values less than or equal to 0.6 are counted as a negative or 'safe' classification, values above are counted as a positive or 'dangerous' classification.

### 4.5 Results

Figure 4 shows the false positive and false negative rates from the experiments. The rates are calculated by comparing the classification from the algorithm with the theoretical classification. Each point on the chart shows the misclassified antigen rate from the beginning of the experiment up to the time indicated on the x-axis. Each series is the average classification error from three runs with the specified migration rate.

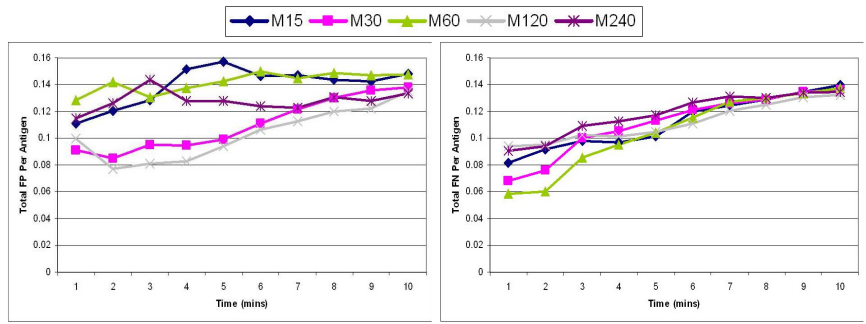

Fig. 4. The classification error rates from the experiments. The false positive rate is shown on the left and the false negative rate is shown on the right

### 4.6 Analysis

The classification error rates rise throughout the experiments. Analysis of the robot's telemetry showed that the error in localisation from the dead-reckoning algorithm was drifting over time. As the measure used to assess the robot's performance relies upon the location of the robot, it is theorized that the classification errors from the first 1-2 minutes are closer to the "true" classification errors, as they will not be as significantly affected by the localisation drift. The use of a theoretical model as a baseline for the experiment could also introduce a constant error offset as the model may not be totally accurate. However, the performance of the algorithm is still high. The highest recorded rate of classification error for the entire experiment is a 0.16 false-positive rate, for the M240 experiment. The higher amount of antigen absorbed before migration increases the occurrence of cases when a DC collects both dangerous and safe antigen, making attributing 'blame' more difficult. The false positive rates all start below 0.14. M30 demonstrated the best performance overall, and appears to give the optimum performance for this particular experiment. It is intuitive to see higher error rates from experiment M15 as a low migration threshold will cause

DC's to migrate after only sampling a small amount of signal. This would result in the classifier being more prone to noise within the system. M120 has amongst the highest false negative rates and the lowest false positive rates. More work will have to be done to understand why this should be the case. One potential cause may be the high range of possible migration thresholds with a tolerance of ±60. M60 yields the lowest false negative rate, but one of the highest false positive rates. The rates of error presented in 4 are lower than expected for this problem, indicating that the DCA is suitable for some robotic applications.

## 5 Conclusions

The misclassifications caused by the dead reckoning errors lead to the results being difficult to judge against the chosen metric. This could be a problem for future applications, as antigen generation for this application is intrinsically location specific. It is proposed that this issue could be resolved by adding a more advanced localisation algorithm based on using sensor readings to compensate for the integration errors.

It has been shown that it is possible to implement the DCA on a real robotic system. Whilst the problem was trivial, the low false positive and false negative rates are promising, especially considering that very little tuning or training has been performed. The implementation did not require any processing to be shared by another machine, so the DCA is scalable for an n-robot system and is usable in circumstances when the robot enters a region with poor communications coverage.

The next intended step for this project is to apply the DCA to a harder classification problem and compare its results with a fuzzy or neural classifier. This will provide an insight into the general performance of the DCA as a robotic classifier. Extending this work to a security system will require two key steps. Firstly the DCA will need to be modified to handle vector antigen instead of integer antigen. This will allow a more extendible representation of the environment to be used by the classifier. Secondly, the signal sources for a security system will need to be more complex than those used for this experiment. A possible source for the PAMP signal would be the error from a trained, non-linear model, correlating physical position to a normal scenario. Large error rates would imply an anomalous situation. The safe and danger signals could be controlled by the robot's physical location and the time of day. It would be advantageous to make the robot less sensitive during office hours and around busy public areas and more sensitive out of office hours and around high-security areas. A more generic anomaly detection system could be achieved through the introduction of a training-data based algorithm. The error rate between what the trained system expects to see and what its current sensor readings tell it, could be used as a source for PAMP signals.

Ultimately a multi-robot system, with dynamically changing routes and shared anomaly information could be developed, each using a DCA to assess the threat level for a given location.

## Acknowledgements

Many thanks to William Wilson for his input to the software architecture and to Daniel Bardsley for his advice on image processing. The authors are very grateful to Mark Hammonds for generating the vector graphics for this paper. This work is financially supported by MobileRobots Inc.

## References

1. SIA: The security industry authority annual report and accounts. Available at http://www.the-sia.org.uk/ (2005-2006)
2. Tickner, A.H., Poulton, E.: Monitoring up to 16 synthetic television pictures showing a great deal of movement,. Ergonomics 14(4) (1973)
3. Greensmith, J., Aickelin, U., Twycross, J.: Articulation and clarification of the dendritic cell algorithm. In: ICARIS'06. (2006)
4. Brooks, R.A.: A robust layered control system for a mobile robot. IEEE J. Robotics and Automation (1986) 14–23
5. Brooks, R.A.: Elephants don't play chess. Robotics and Autonomous Systems 6 (1990) 3–15
6. Castelnovi, M., Miozzo, M., Scalzo, A., Piaggio, M., Sgorbissa, A., Zaccaria, R.: Surveillance robotics: analysing scenes by colours analysis and clustering. In: CIRA. (2003)
7. Everett, H., Gilbreath, G., Heath-Pastore, T., Laird, R.: Controlling multiple security robots in a warehouse environment. In: AIAA/NASA Conference on Intelligent Robots. (1994)
8. Pastore, T., Everett, H., Bonner, K.: Mobile robots for outdoor security applications. In: ANS'99. (1999)
9. Massios, N., Voorbraak, F.: Hierarchical decision-theoretic robotic surveillance. In: IJCAI'99 Workshop on Reasoning with Uncertainty in Robot Navigation. (1999)
10. Greensmith, J., Aickelin, U., Cayzer, S.: Introducing dendritic cells as a novel immune inspired algorithm for anomaly detection. In: ICARIS'05. (2005)
11. Lutz, M., Schuler, G.: Immature, semi-mature and fully mature dendritic cells: which signals induce tolerance or immunity? Trends in Immunology 23(9) (2002)
12. Greensmith, J., Twycross, J., Aickelin, U.: Dendritic cells for anomaly detection. In: Congress on Evolutionary Computation (CEC). (2006)
13. J. Kim, P. J. Bentley, C.W.M.A., Hailes, S.: Danger is ubiquitous: Detecting misbehaving nodes in sensor networks using the dendritic cell algorithm. In: ICARIS '06. (2006)
14. Neal, M., Feyereisl, J., Rascuna, R., Wang, X.: Don't touch me, I'm fine: Robot autonomy using an artificial innate immune system. In: ICARIS'06. (2006)
15. Swain, M., Ballard, D.: Color indexing. International Journal of Computer Vision 7(1) (1991)
16. Krautmacher, M., Dilger, W.: AIS based robot navigation in a rescue scenario. In: LNCS, Artificial Immune Systems. (2004)
17. Twycross, J., Aickelin, U.: Libtissue - implementing innate immunity. In: Congress on Evolutionary Computation (CEC'06). (2006)